\documentclass[]{spie}  

\usepackage{amsmath,amsfonts,amssymb}
\usepackage{subfig}
\usepackage{graphicx}
\usepackage{booktabs}
\usepackage[colorlinks=true, allcolors=blue]{hyperref}
\usepackage{cleveref}
\usepackage{algorithm}
\usepackage{algpseudocodex}
\usepackage{upgreek}

\graphicspath{{./figs/}}

\newcommand{\etal}{\textit{et al}. }

\newcommand{\eg}{\textit{e}.\textit{g}.}

\title{GPU-Accelerated Matrix Cover Algorithm for Multiple Patterning Layout Decomposition}

\author[1]{Guojin Chen}
\author[2]{Haoyu Yang}
\author[1]{Bei Yu}

\affil[1]{The Chinese University of Hong Kong, NT, Hong Kong}
\affil[2]{NVIDIA Crop.}

\begin{document}
\maketitle

\begin{abstract}
    Multiple patterning lithography (MPL) is regarded as one of the most promising ways of overcoming the resolution limitations of conventional optical lithography due to the delay of next-generation lithography technology.
    As the feature size continues to decrease, layout decomposition for multiple patterning lithography (MPLD) technology is becoming increasingly crucial for improving the manufacturability in advanced nodes.
    The decomposition process refers to assigning the layout features to different mask layers according to the design rules and density requirements.
    When the number of masks $k \geq 3$, the MPLD problems are $\mathcal{NP}$-hard and thus may suffer from runtime overhead for practical designs.
    However, the number of layout patterns is increasing exponentially in industrial layouts, which hinders the runtime performance of MPLD models.
    In this research, we substitute the CPU's dance link data structure with parallel GPU matrix operations to accelerate the solution for exact cover-based MPLD algorithms.
    Experimental results demonstrate that our system is capable of full-scale, lightning-fast layout decomposition,
    which can achieve more than 10$\times$ speed-up without quality degradation compared to state-of-the-art layout decomposition methods.
\end{abstract}

\section{Introduction}
\label{sec:intro}  

The general process of MPLD is to assign the layout features close to each other into the different masks to enhance the lithography resolution
since the features are far away enough to be printed with the existing lithography techniques.
Previous literature generally models the MPLD to graph coloring problems and solves them using linear programming (LP) methods or variants.
Unlike the classical coloring problems, the MPLD problems have their features.
1) The graph node representing the layout polygon can be split into multiple polygon segments, called stitch.
2) There are other rules beyond the wildly adopted spacing constraint for the same color.
Those constraints impose different challenges to the MPLD problems.
Typically, MPLDs are formulated into mathematical optimization models, which can be roughly categorized into three types~\cite{openmpl}:
1) integer linear programming (ILP) and its relaxation, 2) graph-based methods, 3) search-based approaches.
Specifically, ILP for double patterning layout decomposition (DPLD)~\cite{DPL-ICCAD2009-Xu,DPL-TCAD2010-Kahng,DPL-TCAD2010-Yuan,SADP-ASPDAC2014-Gao}
or triple patterning layout decomposition (TPLD)~\cite{TPL-TCAD2015-Yu,TPL-ICCAD2013-Yu,TPLEC-JM3-2015-Yu, TPL-JM3-2017-Lin,TPL-SPIE2016-Lin,TPL-ICCAD2011-Yu,TPLEC-SPIE2013-Yu,TPL-DAC2014-Yu,TPL-SPIE2012-Lucas,TPL-SPIE2014-Yu}.
The relaxation techniques for ILP methods are well-researched due to the $\mathcal{NP}$-hardness of TPLD and QPLD~\cite{TPL-TCAD2015-Yu,TPL-JM3-2017-Lin,TPL-TC2017-Li,MPL-DAC2015-Pan}.
The graph feature of the input layout makes it natural to handle the MPLD with graph-theoretical algorithms,
\eg, the maximal-independent set (MIS) \cite{TPL-DAC2012-Fang}, the shortest path \cite{TPL-ICCAD2012-Tian}.
Another category is to use search-based algorithms following the divide-and-conquer principle,
performing the search procedure on the sub-graphs~\cite{TPL-DAC2013-Kuang,TPL-DAC2016-Chang}.
Nevertheless, graph coloring-based models encounter substantial obstacles when sophisticated rules are required.
To address the intricate rules and density balance, exact cover(EC)-based MPLD models have been suggested~\cite{TPL-TCAD2017-Jiang}.



Donald Knuth invented Algorithm $\text{X}^{\ast}$ to solve the EC problem
and further suggested an efficient implementation technique called dancing links (DLX)~\cite{knuth2000dancing},
using doubly-linked circular lists to represent the matrix of the problem.
Jiang and Chang~\cite{TPL-TCAD2017-Jiang} designed a general and flexible MPLD framework based on augmenting DLX with LD task-related treatments,
which can concurrently consider complex coloring rules and maintain density balancing.
Li \etal presented OpenMPL~\cite{openmpl}, an open-source LD framework,
which introduced an improved flexible EC-based algorithm achieving better quality with a sacrifice in the runtime.
However, this sacrifice in runtime is even more evident in the larger industrial designs, making the EC-based methods less practical when compared with ILP-based methods.
With the increasing power of modern graphics processing units (GPUs), numerous successful applications use GPU to
accelerate the design automation tasks, \eg, mask optimization\cite{OPC-ICCAD2021-DevelSet,OPC-DATE2021-Yu,OPC-ICCAD2020-DAMO,MCH-DATE2022-Yang,AdaOPC-iccad-zhao,OPC-TCAD2020-Geng}, design space exploration \cite{PTPT-TCAD22-Geng,HS-TCAD22-Geng,BSL-TNNLS22-CHEN,HSD-TCAD2022-Geng}, and layout generation~\cite{LayouTransformer-iccad22-wen,deepattern-dac19-yang,pc-ispd21-li}.
Recent work pioneers using GPU and deep learning on pre or-post process of MPLD~\cite{OPC-ICCAD2017-Ma,OPC-DAC2020-Zhong,TPL-DAC2020-Li,MPL-VLSI-SOC2017-Ma}.
However, the acceleration of the Layout decomposition algorithm itself has not been explored.
It's necessary to explore the GPU-accelerated EC-based methods for large industrial designs.
In this paper, we propose a GPU-accelerated matrix cover algorithm for MPLD.
Our main contributions are:
\begin{itemize}
  \item We replace the CPU-based dancing link algorithms by leveraging the CUDA indexing model to solve the EC-based MPLD problems.
  \item We apply task parallelism to decompose the layout graph and improve the computation efficiency by parallel execution of CUDA kernel functions.
  \item We develop our GPU acceleration MPLD algorithms on top of the open-sourced decomposer OpenMPL~\cite{openmpl} to ensure usability and scalability.
\end{itemize}


\begin{figure}[tb!]
  \centering
    \subfloat {\includegraphics[width=.24\linewidth]{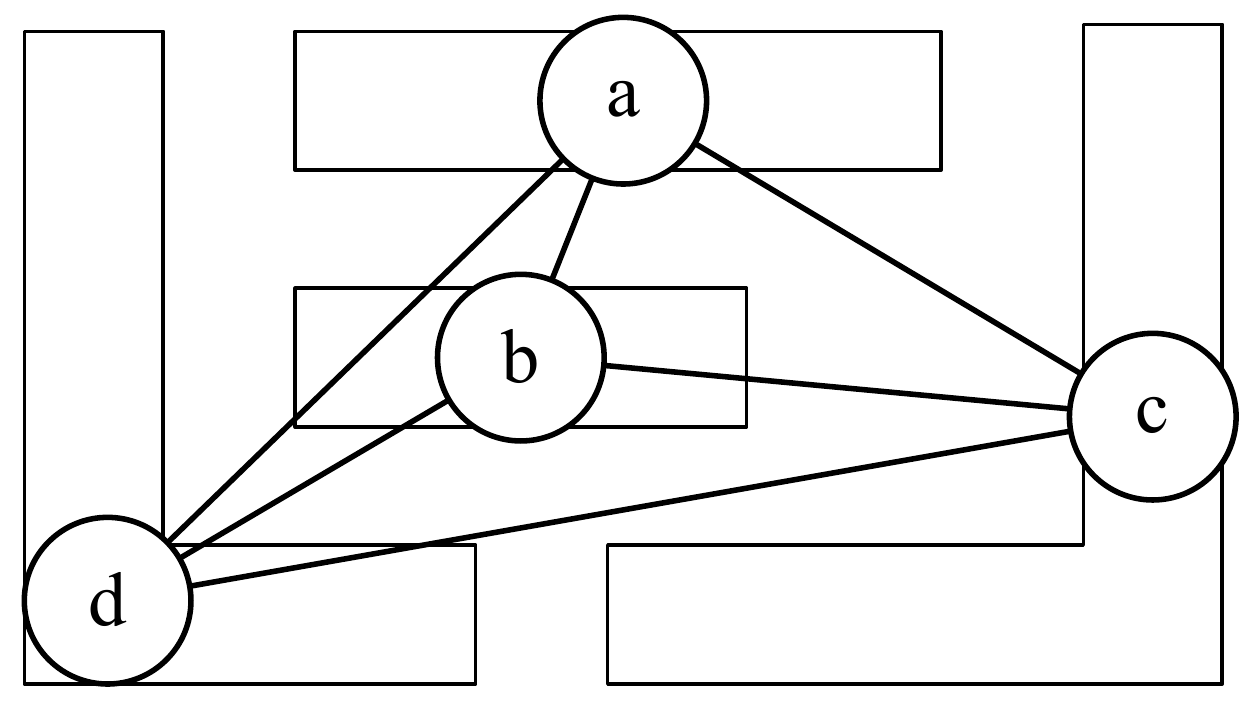}} \hspace{5em}
    \subfloat {\includegraphics[width=.24\linewidth]{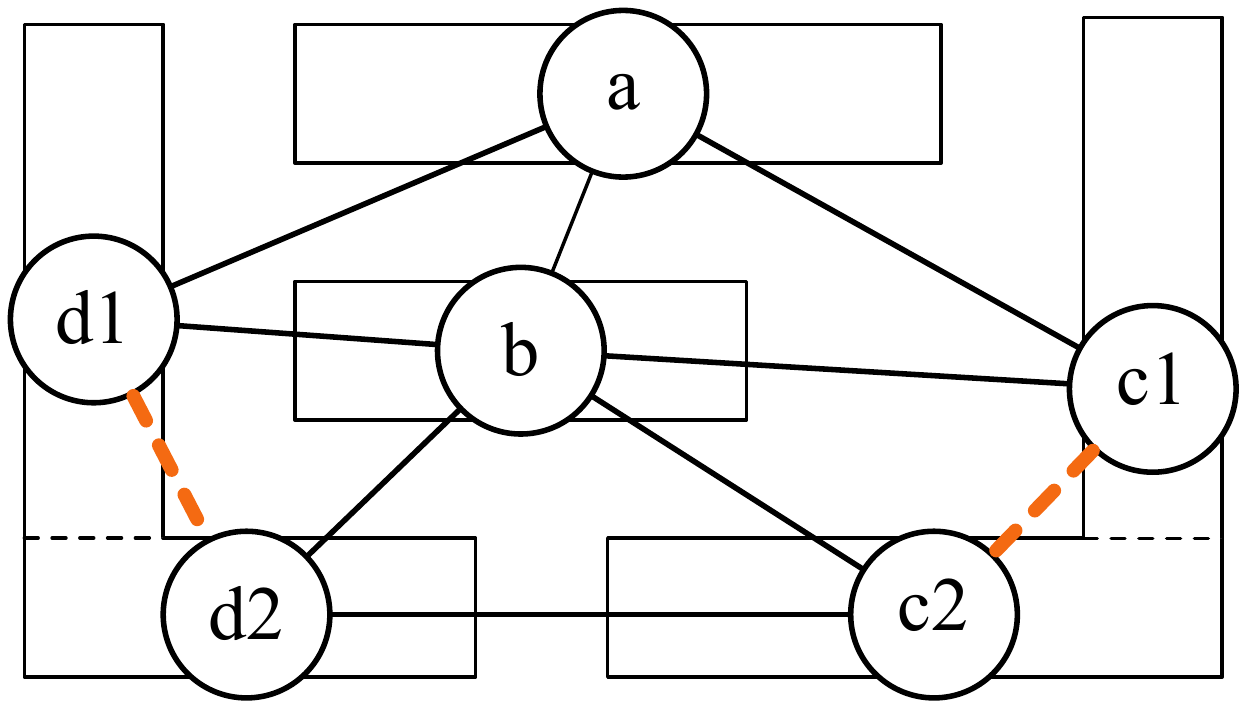}} \\ \vspace{.5em}
    \subfloat {\includegraphics[width=.24\linewidth]{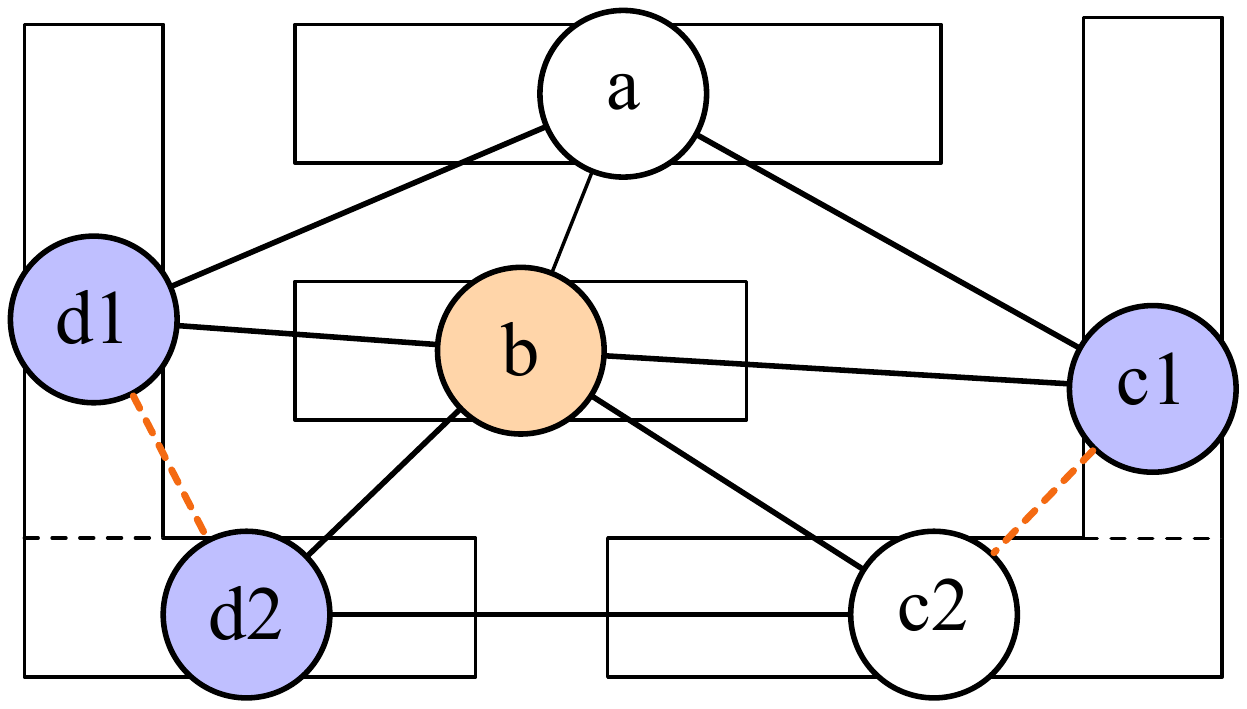}} \hspace{5em}
    \subfloat {\includegraphics[width=.24\linewidth]{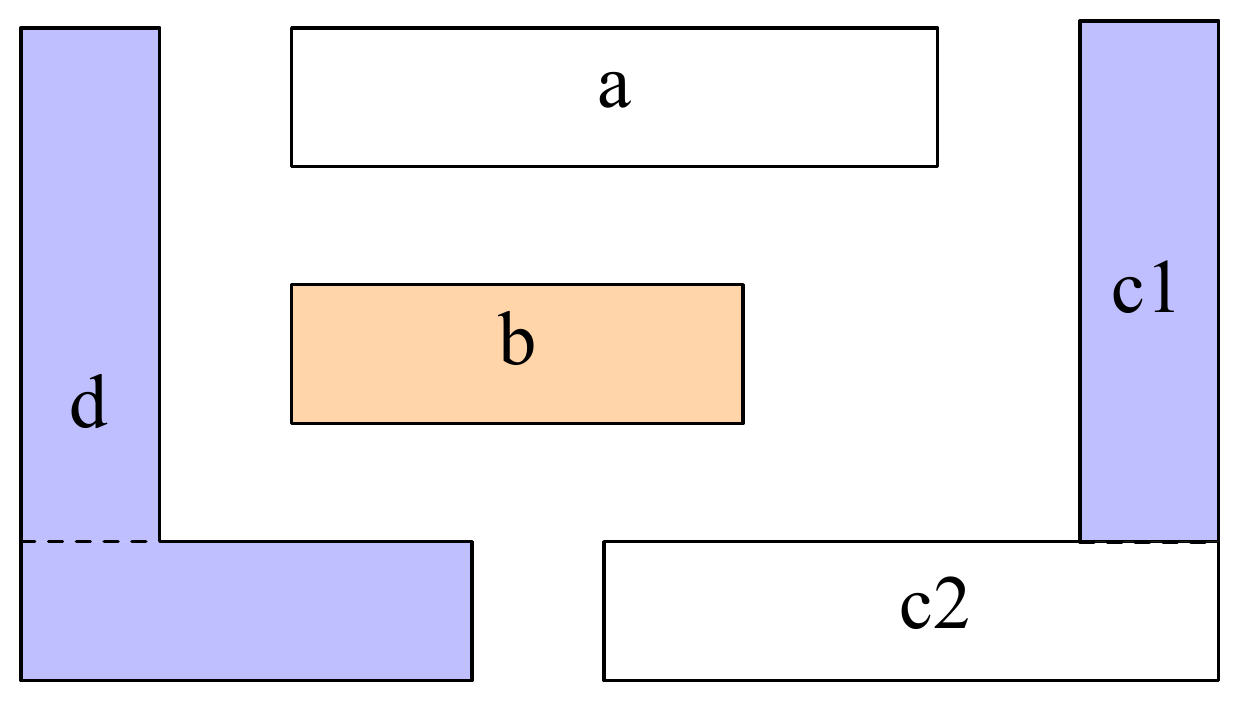}} \vspace{.4em}
    \caption{
      An example of TPLD with stitches.
      (a) The input layout consists of multiple polygons.
      (b) The layout graph (LG) construction process. The LG is 4-clique therefore not 3-colorable.
      (c) Stitches insertion process. Insert two stitches into the original LG to eliminate the 4-clique.
      (d) Decomposition process. The LG will be decomposed using a coloring solver with stitch candidate generation.
      The final layout will be decomposed into three masks colored with different colors.
    }
  \label{fig:stitch}
\end{figure}

\section{Algorithm}
\label{sec:algo}
The detailed formulation of the MPLD problem is introduced in \Cref{subsec:mpld_prob}.
Our algorithm is built on top of an open-sourced layout decomposer OpenMPL~\cite{openmpl},
which implements various CPU-based MPLD methods. The decomposition flow is introduced in \Cref{subsec:mpld_flow}.
As proposed in \Cref{subsec:gpu_mpld} We replace the decomposition solver with our GPU-accelerated matrix cover algorithm.

\subsection{Problem formulation}
\label{subsec:mpld_prob}

As depicted in \Cref{fig:stitch}, given an input layout specified by features in polygonal shapes, the layout can be translated into
an undirected layout graph $G=\{V,E\}$, where every node $v_i \in V$ corresponds to one feature in layout and each edge $e_{ij} \in E$ is used to characterize the relationships between features. 
Considering conflict and stitch relationships, $E$ is composed of these two kinds of edges, denoted by
$E=\{CE \cup SE\}$, where $SE$ is the set of stitch edges and $CE$ is the set of conflict edges.
The MPLD problem can be formulated as below:
\begin{subequations}  \label{formula:mpl}
    \begin{align}
        \underset{\vec{x}}{\min} \ \ & \sum   c_{ij} + \alpha\sum s_{ij},   \\
        \textrm{s.t.} \ \
        & c_{ij} = (x_i==x_j),          && \forall e_{ij} \in CE, \\
        & s_{ij} = (x_i \neq x_j),      && \forall e_{ij} \in SE, \\
        & x_i \in \{0, 1, \ldots ,k\},  && \forall x_i \in \vec{x},
    \end{align}
\end{subequations}
where $x_{i}$ is a variable for the $k$ available colors of the pattern $v_{i}$, $c_{ij}$ is a binary variable representing conflict edge $e_{ij}\in CE$,
$s_{ij}$ stands for stitch edge $e_{ij}\in SE$,
$\alpha$ is a user-defined parameter and is set as 0.1 by default in our framework to assign relative importance between the conflict cost and the stitch cost.
If two nodes, $x_{i}$ and $x_{j}$, within the minimal coloring distance are assigned the same color (i.e.~$x_i==x_j$), then $c_{ij}=1$.
On the contrary, $s_{ij}=1$ when two nodes connected by stitch edge are assigned a different color (i.e.~$x_i \neq x_j$).
The objective function is to minimize the weighted summation of the conflict number and the stitch number.

\subsection{The MPLD flow}
\label{subsec:mpld_flow}
\begin{figure}[tb!]
  \centering
  \includegraphics[width=.8\linewidth]{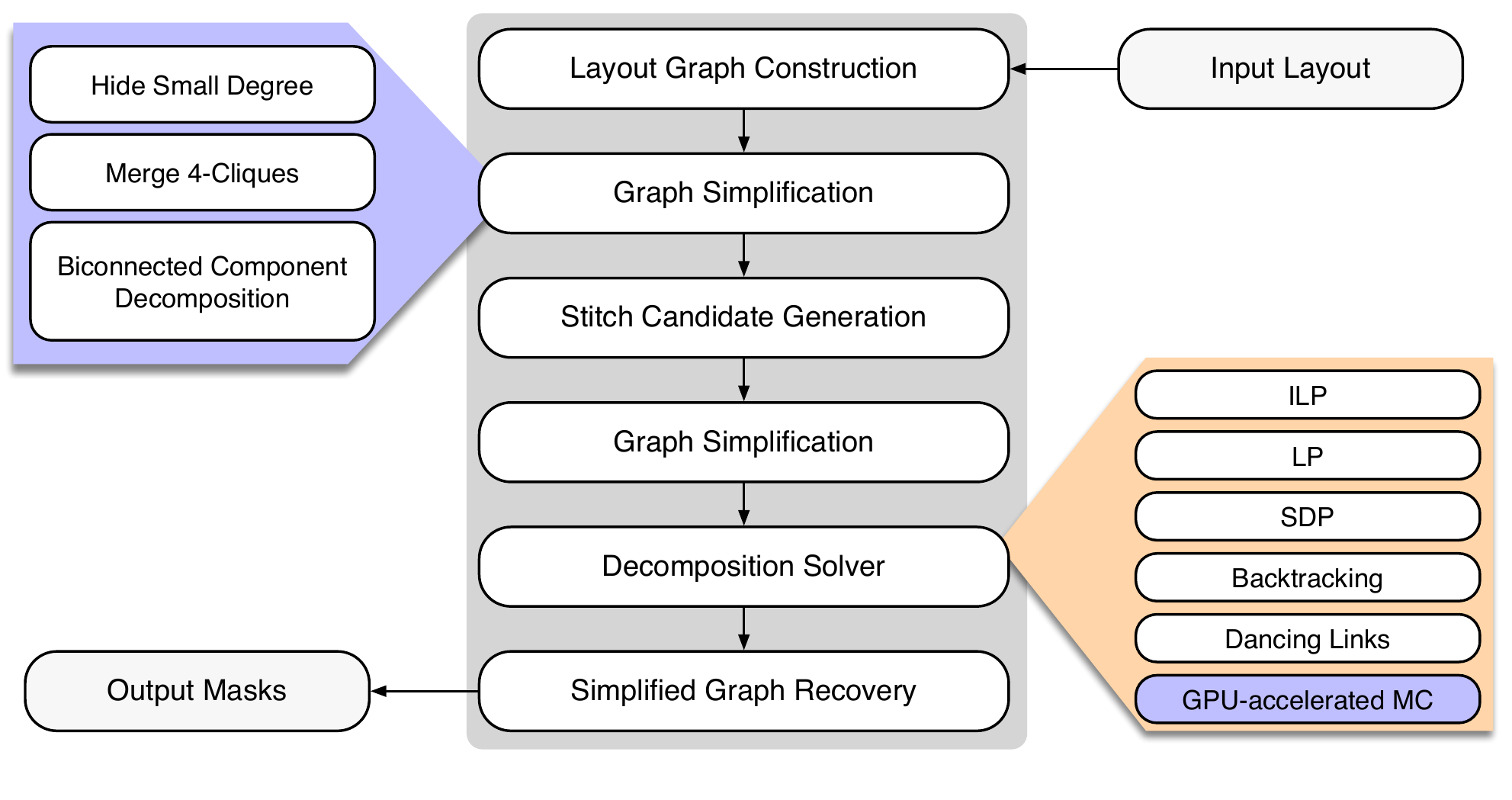}
  \caption{The overall flow for our framework, the coloring solver is replaced with our GPU-accelerated matrix cover solver.}
  \label{fig:flow}
\end{figure}

As illustrated in \Cref{fig:flow}, we utilize a part of OpenMPL flow.
The chip layout will be first transformed into a layout graph (LG) by a vector of rectangle pointers.
The second step is to simplify the layout graph with stitch insertion, after which the layout graph will be decomposed into decomposed graph (DG) with stitches.
Then we can call the graph coloring solver to solve the MPLD problems.
As illustrated in \Cref{fig:flow}, in this step, we replace the EC-based solver with our GPU-accelerated matrix cover solver.
Finally, the framework will recover the nodes removed in simplification step and assigns the coloring results from the coloring solver.

\subsection{GPU-accelerated matrix cover for MPLD}
\label{subsec:gpu_mpld}
\begin{figure}[tb!]
  \centering
  \includegraphics[width=.55\linewidth]{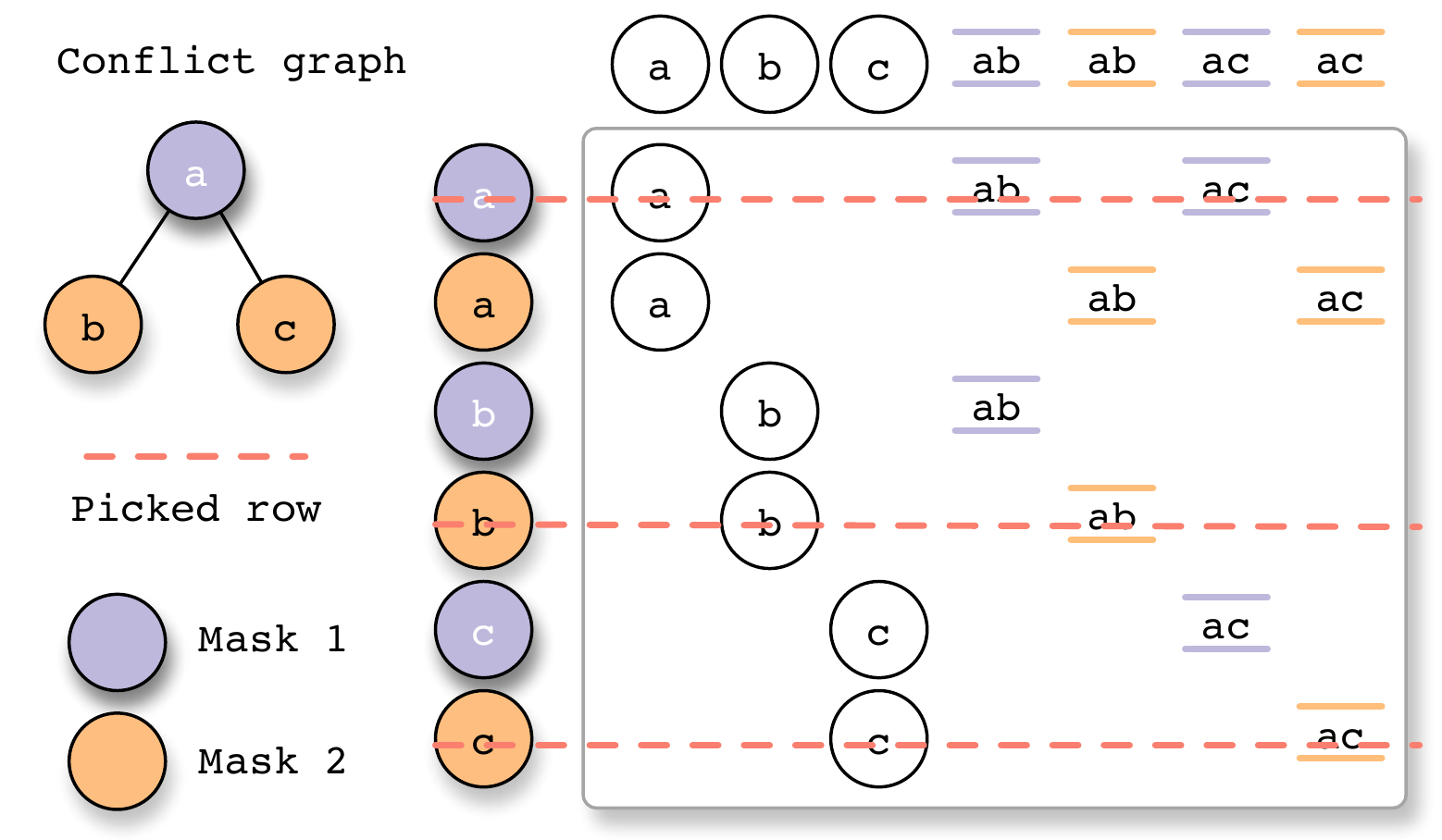}
  \caption{
    Exact cover matrix for double patterning layout decomposition with basic coloring rule.
  }
  \label{fig:dancing_links}
\end{figure}

The original EC-based algorithm~\cite{TPL-TCAD2017-Jiang} models the MPLD as a matrix cover problem, as depicted in \Cref{fig:dancing_links}.
The input of the algorithm is a no-stitch graph $G_p = \{V_p, E_p\}$.
Since $G_p$ tends to be spare, Knuth~\cite{knuth2000dancing} suggested using DLX to solve it efficiently.
The original layout will be transformed into a homogeneous graph and further translated into a binary matrix of ``0''s and ``1''s.
Then the solution for the MPLD problem is transformed into solving the EC problem,
which means finding a set of rows containing exactly one ``1'' in each column (\Cref{fig:dancing_links}).
The cover and uncover operations can be transformed into the removal and recovery of a doubly linked list,
\begin{equation}
  \text{Cover}: L[R[x]] \rightarrow L[x], R[L[x]] \rightarrow R[x],\ \  \text{Uncover}: L[R[x]] \rightarrow x, R[L[x]] \rightarrow x
\end{equation}
where $L[x]$ and $R[x]$ point to the left and the right node of the linked list.

\begin{figure}[tb!]
  \centering
  \includegraphics[width=.78\linewidth]{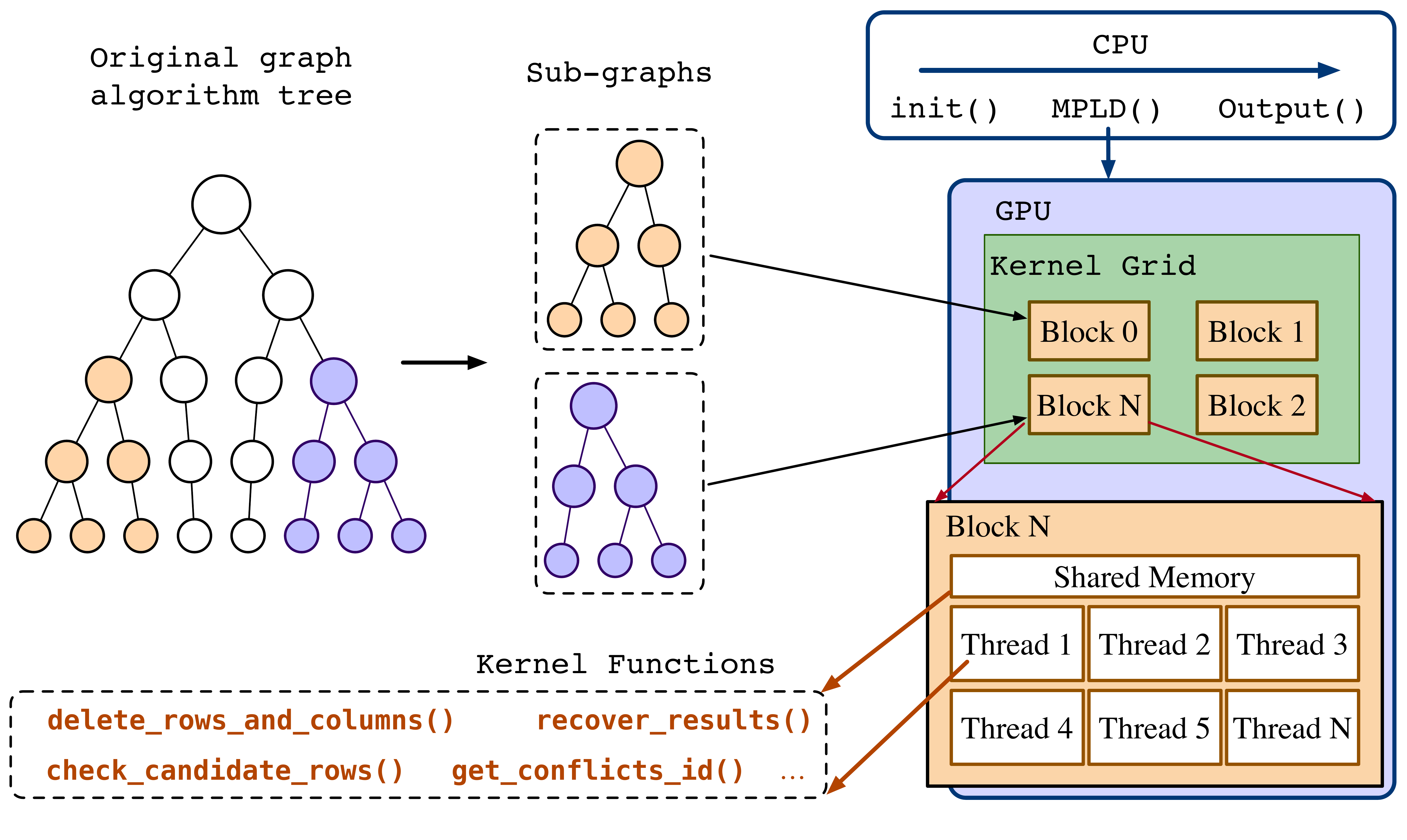}
  \caption{
    The hierarchical GPU acceleration model of our matrix cover algorithm.
  }
  \label{fig:mtxcover_gpu}
\end{figure}

The hierarchical acceleration of GPU-based matrix cover solver is illustrated in \Cref{fig:mtxcover_gpu} and \Cref{alg:gpu-mpld}.
The three-dimensional indexing of CUDA programming model provides a natural way to index the matrix elements thus the DLX algorithm can be accelerated by GPU with the index $x$ set as:
\begin{equation}
  x = blockIdx.x \times blockNum + threadIdx.y
\end{equation}
where the $blockIdx$ and the $threadIdx$ are the indexer of CUDA programming model, the $blockNum$ is a pre-defined hyperparameter indicating the degree of parallelism.
As shown in \Cref{alg:gpu-mpld}, the original DG will be decomposed into sub-graphs for acceleration (line 1 - line 3).
Then the sub-graphs will be parallelly executed on different blocks (line 4) of a GPU to accelerate the DLX algorithm (line 4 - line 18).
The \textit{cover} and \textit{uncover} operations in line 8 and line 16 are implemented as CUDA kernel functions \eg, \texttt{delete\_rows\_and\_columns}, \texttt{recover\_results} and so on.
As depicted in \Cref{fig:mtxcover_gpu},
the kernel functions will be organized under the indexing and perform DLX in parallelization.
Moreover, the shared memory of GPU is also used for global communication of different threads for further acceleration.

\begin{algorithm}[h]
  \caption{GPU-accelerated matrix cover algorithm}
    \begin{algorithmic}[1]
        \Require Layout graph $G$ and converted exact cover matrix $M$
        \Ensure Colored graph and conflict features
        \State $graphID = blockIdx.x$
        \State $subgraphID = threadIdx.y$
        \State $idx = blockIdx.x * graph\_per\_block + subgraphID$
        \For{each subgraph}
          \If{ no column of $idx$ remains or all columns of $M$ are covered}
              \State Return \Comment{The solution has been found.}
          \EndIf
          \State Select an uncovered column $cl$ with only one related row or in BFS order of $G$
          \State Cover $cl(idx)$
          \If{no $cl$'s related rows remain}
              \State Mark $(cl, cl')$ as one conflict candidate
              \State where $cl$'s is the column that has covered final related row of $cl$
          \EndIf
          \For{each row $rw$ related to $cl$}
              \State Include $rw$ into the current solution
              \State Cover $rw$ and its affected rows
              \State Call GPU-accelerated matrix cover with $G$ and $M$
              \State Uncover $rw$ and its affected rows
              \State Exclude $rw$ from the current solution
          \EndFor
          \State Uncover $cl$
        \EndFor
        \State Return
    \end{algorithmic}
  \label{alg:gpu-mpld}
\end{algorithm}

\section{Experimental results}

The framework is implemented in C++ on an Intel Core 2.9-GHz Linux machine with Nvidia GeForce RTX 2080 GPU and nvcc 11.0 compiler.
Regarding kernel execution, we assign 32 threads per block with one block for each sub-graph.
We compare our results with the original EC~\cite{TPL-TCAD2017-Jiang}, OpenMPL EC~\cite{openmpl} on ISCAS benchmarks.
The ISCAS benchmarks are widely used in previous works. The minimum coloring spacing is set to 120 nm for the first ten cases and 100 nm for the last five cases, which are the same settings as \cite{TPL-TCAD2017-Jiang}, and \cite{openmpl}.
We show the results in \Cref{tab:results}, where the ``time(s)'' column is the total simplification and decomposition time of graphs which have redundant stitches to be removed.
The columns ``st\#'' and ``cn\#'' are the stitch and the conflict numbers.
Compared with original EC, our GPU-accelerated solver can achieve 17.6 $\times$ runtime speed-up with 2\% fewer stitches.
Compared with OpenMPL EC, our algorithm can achieve 20 $\times$ speed-up and 1\% fewer stitches with reasonable sacrifice on conflict number.
\begin{table}[tb!]
  \centering
  \caption{Result comparison with the state-of-the-art}
  \label{tab:results}
    \begin{tabular}{c|cc|ccc|ccc|ccc}
    \toprule
              & \multicolumn{2}{c|}{Graph Info}        & \multicolumn{3}{c|}{Original EC} & \multicolumn{3}{c|}{OpenMPL EC} & \multicolumn{3}{c}{Ours} \\
      Circuit & $\lvert V \rvert$ & $\lvert V \rvert$ & time(s) & st$^\#$ &cn$^\#$ & time(s) &st$^\#$  &cn$^\#$ & time(s)  &st$^\#$ &cn$^\#$ \\ \midrule
      c432    & 1109              & 1222              & 0.005   &4        &0       & 0.008   &4        &0       & 0.000367 &4       &0       \\
      c499    & 2216              & 2817              & 0.004   &0        &0       & 0.006   &0        &0       & 0.000044 &0       &0       \\
      C880    & 2411              & 2686              & 0.005   &7        &0       & 0.007   &7        &0       & 0.000368 &7       &0       \\
      C1355   & 3262              & 3326              & 0.007   &3        &0       & 0.018   &3        &0       & 0.000767 &3       &0       \\
      C1908   & 5125              & 5598              & 0.008   &1        &0       & 0.022   &1        &0       & 0.001019 &1       &0       \\
      C2670   & 7933              & 9336              & 0.014   &6        &0       & 0.021   &6        &0       & 0.00085  &6       &0       \\
      C3540   & 10189             & 11968             & 0.029   &8        &1       & 0.035   &8        &1       & 0.006474 &8       &1       \\
      C5315   & 14603             & 16881             & 0.019   &9        &0       & 0.033   &9        &0       & 0.000511 &9       &0       \\
      C6288   & 14575             & 15605             & 0.114   &203      &8       & 0.142   &204      &1       & 0.007601 &203     &8       \\
      C7552   & 21253             & 24372             & 0.028   &21       &1       & 0.055   &21       &1       & 0.003175 &21      &1       \\
      S1488   & 4611              & 5504              & 0.008   &2        &0       & 0.007   &2        &0       & 0.000451 &2       &0       \\
      S38417  & 67696             & 79527             & 0.127   &54       &19      & 0.175   &54       &19      & 0.005955 &54      &19      \\
      S35932  & 157455            & 186052            & 0.286   &48       &44      & 0.299   &40       &44      & 0.018273 &40      &44      \\
      S38584  & 168319            & 196072            & 0.291   &117      &36      & 0.323   &117      &36      & 0.009688 &117     &36      \\
      S15850  & 159952            & 190796            & 0.285   &100      &34      & 0.342   &100      &34      & 0.014451 &100     &34      \\ \midrule
      average & -                 & -                 & 0.082   &38.9     &9.5     & 0.1     &38.4     &9.07    & 0.004666 &38.3    &9.5     \\
      ratio   & -                 & -                 & 1       &1        &1       & 1.2     &0.99     &0.95    & \textbf{0.0569}   &0.98 &1 \\ \bottomrule
      \end{tabular}
\end{table}

\begin{figure}[tb!]
  \centering
  \subfloat[]{\includegraphics[width=.42\linewidth]{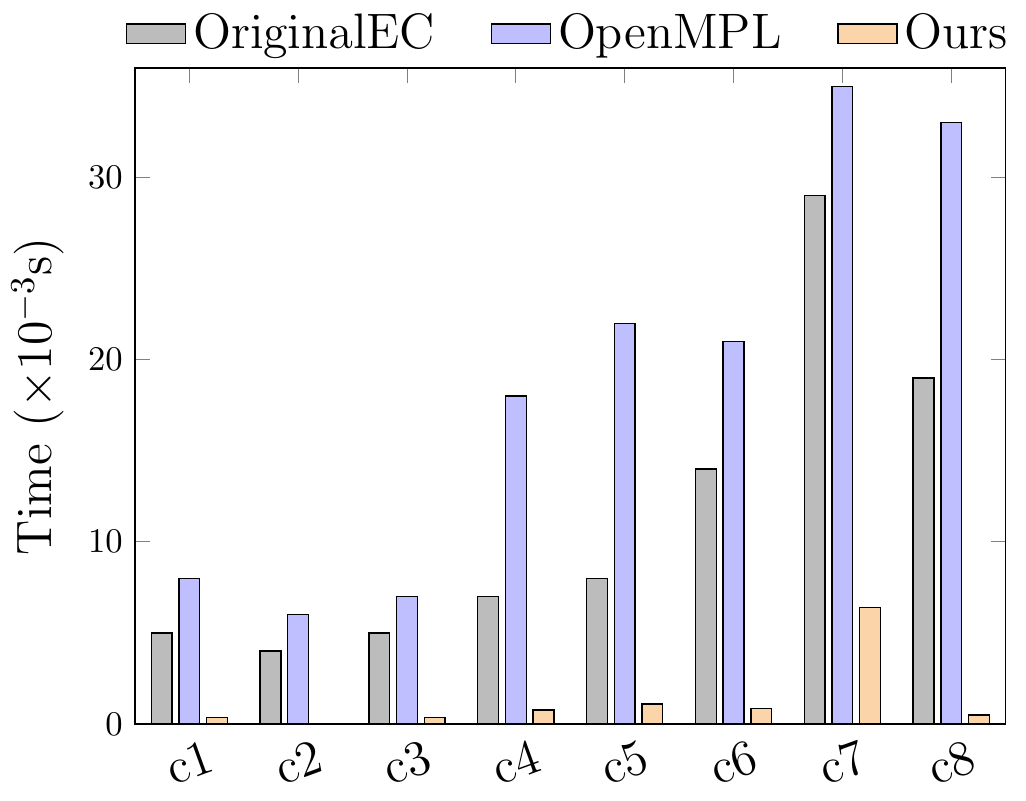} \label{fig:bar-graph}} \hspace{2em}
  \subfloat[]{\includegraphics[width=.42\linewidth]{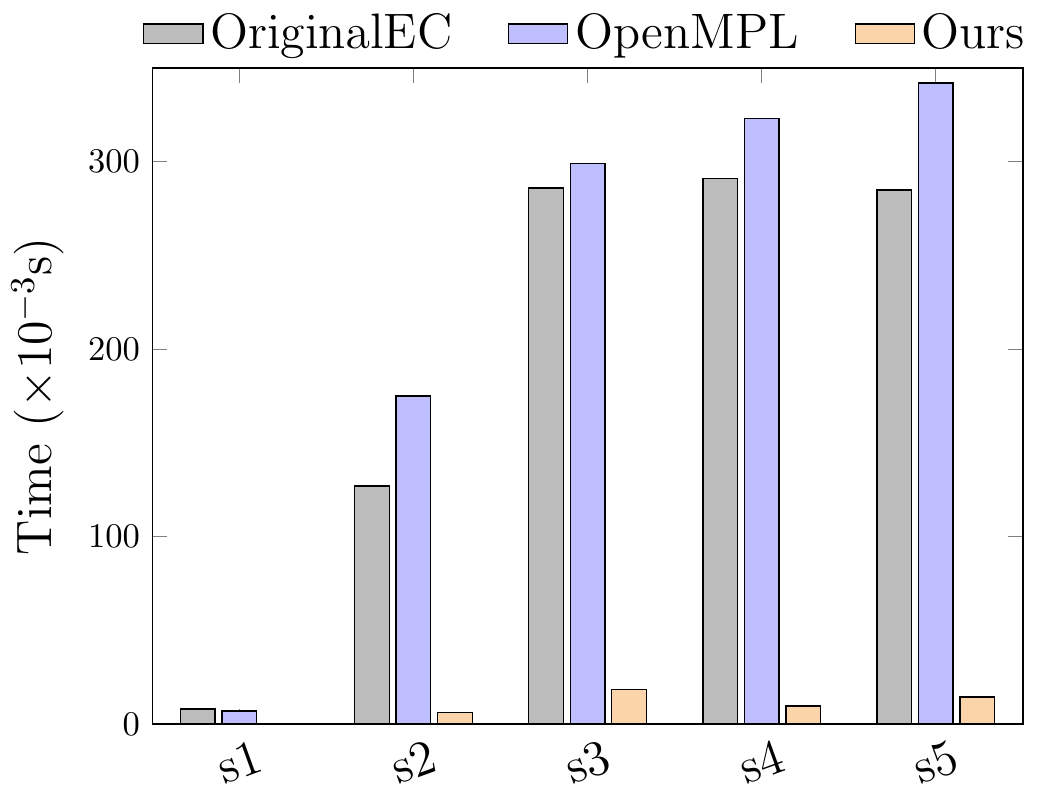} \label{fig:bar-large}}
  \caption{
    (a) The result comparisons on first ten smaller cases of ISCAS benchmarks.
    (a) The result comparisons on last five larger cases of ISCAS benchmarks.
  }
\end{figure}

\section{Conclusion}

In this paper, we propose a GPU-accelerated matrix cover algorithm for multiple patterning layout decomposition problems.
Then we develop a set of GPU-efficient data structures and algorithms to accelerate the coloring process.
We leverage the CUDA programming model to implement the DLX algorithm on GPU in parallelization.
Compared with the state-of-the-art EC engine, our GPU-accelerated algorithm can achieve up to 17.6 $\times$ speed-up on large designs.

\bibliographystyle{spiebib} 
\bibliography{ref/top.bib,./ref/GPU.bib,./ref/MPL.bib} 

\end{document}